\documentclass[submission,copyright,creativecommons]{eptcs}
	\providecommand{\event}{DICE-FOPARA 2019}
\usepackage{newtxtext,newtxmath}
\usepackage{url}
\usepackage{graphicx}

\usepackage{amssymb}
\usepackage{amsmath}
\usepackage{xspace}
\usepackage{lineno}
\usepackage{tabularx}
\usepackage{microtype}
\usepackage{verbatim}
\usepackage{xcolor}
\usepackage{enumitem}
\usepackage{lineno}
\usepackage{todonotes}
\usepackage[normalem]{ulem}

\setlist{nosep}
\AtBeginDocument{
\setlength{\abovedisplayskip}{3pt plus 0.5pt minus 0.5pt}
\setlength{\belowdisplayskip}{3pt plus 0.5pt minus 0.5pt}
\setlength{\abovedisplayshortskip}{1pt plus 0.5pt minus 0.5pt}
\setlength{\belowdisplayshortskip}{3pt plus 0.5pt minus 0.5pt}
}
\newcommand{\To}{\Rightarrow}

\newcommand{\PROP}{\ensuremath{\mathrm{PROP}}\xspace}
\newcommand{\LIT}{\ensuremath{\mathrm{Lit}}\xspace}

\newcommand{\FACTS}{\ensuremath{F}}
\newcommand{\LAB}{\ensuremath{\mathrm{Lab}}\xspace}

\newcommand{\non}{\ensuremath{\mathord{\sim}}}

\RequirePackage{ifthen}
\newcommand{\set}[2][\relax]{%
 \ifthenelse{\equal{#1}{auto}}{\left\{#2\right\}}{#1\{#2#1\}}%
}

\newtheorem{definition}{Definition}
\newtheorem{scenario}{Scenario}

\title{Applications of Linear Defeasible Logic: combining resource consumption and exceptions to energy management and business processes}
\def\titlerunning{Applications of Linear Defeasible Logic}
\def\authorrunning{Olivieri et al.}
\providecommand{\event}{DICE-FOPARA 2019}

\author{Francesco Olivieri, Guido Governatori
\institute{Data61, CSIRO, Australia}
\and
Claudio Tomazzoli, Matteo Cristani
\institute{University of Verona, Italy}}

\begin{document}

\maketitle

\begin{abstract}

Linear Logic and Defeasible Logic have been adopted to formalise different
features of knowledge representation: consumption of resources, and non monotonic 
reasoning in particular to represent exceptions. Recently, a framework to combine sub-structural features,
corresponding to the consumption of resources, with defeasibility aspects to
handle potentially conflicting information, has been discussed in literature, by some of 
the authors. Two applications emerged that are very relevant: energy management and
business process management. We illustrate a set of guide lines to determine
how to apply linear defeasible logic to those contexts.

\end{abstract}

\section{Introduction}\label{section:intro}
Resource-based logical framework have been employed to accommodate
varieties of methods that propose to introduce relevant limits to
the reasoning process. In particular, in the idealised representation of rational
agents, these agents have unlimited reasoning power, complete knowledge of the environment
and their capabilities, and unlimited resources. Over the years, a few
approaches (using different logics) have been advanced to overcome some of
these ideal (and definitely unrealistic) assumptions. 

In \cite{linear-agent,linear-agent-Mel1,linear-agent-Mel2}, the authors
propose the use of Linear Logic to model the notion of resource utilisation,
and to generate which plans the agent adopts to achieve its goals. In the same
spirit, \cite{jaamas:bio,tplp:goal} address the problem of agents being able
to take decisions from partial, incomplete, and possibly inconsistent
knowledge bases, using (extensions of) Defeasible Logic (a computational and
proof theoretic approach) to non-monotonic reasoning and reasoning with
exceptions. 
While these last two approaches seem very far apart, they are both based on
proof theory (where the key notion is on the idea of (logical) derivation),
and both logics (for different reasons and different techniques) have been
used for modelling business processes
\cite{Kanovich199748,Tanabe1997156,DBLP:journals/apal/EngbergW97,rao2006composition,DBLP:conf/prima/OlivieriGSC13,DBLP:conf/prima/OlivieriCG15,DBLP:conf/edoc/GhooshchiBGOS17}.

Formally, a business process can be understood as a compact representation of
a set of traces, where a trace is a sequence of tasks. A business process is
hence equivalent to a set of plans with possible choices, or a partially specified plan. 
The idea behind the work mentioned above is to allow agents to use their deliberation phase to
determine the business processes (instead of the plans) to execute.

In this paper, we discuss some applications of the combination of linear logic 
and a computationally oriented non-monotonic formalism known as 
defeasible logic. The
\emph{linear defeasible logic}  has been
introduced recently \cite{DBLP:conf/prima/OlivieriGCBT18}.

To provide the above mentioned framework we introduce here a notion of
logic as a rule-based system. We can distinguish rules (or
sequents, or instances of a consequence relation) and inference (or derivation) rules.  A rule
specifies that some consequences follow from some premises, while a derivation rule provides a
recipe to determine the valid steps in a proof or derivation.  A classical example of a derivation
rule is Modus Ponens (i.e., from `$\alpha\rightarrow\beta$' and $\alpha$  to derive $\beta$). A rule
can be understood as a pair 
\[
  \Gamma\vdash \Theta,
\] 
where $\Gamma$ and $\Theta$ are collections of
formulas in an underling language. In Classical Logic, $\Gamma$ and $\Theta$ are sets of formulas
and in Intuitionistic Logic $\Theta$ is a singleton. Thus the rules 
\[
  \alpha,\beta \vdash \gamma,\delta \qquad \text{ and } \qquad
  \beta,\alpha \vdash \delta, \gamma
\] 
are the same rule. Where ``,'' in the antecedent
$\Gamma$ is understood as conjunction and disjunction in the consequent $\Theta$. In substructural
logics (e.g., Lambek Calculus, and the family of Linear Logics), $\Gamma$ and $\Theta$ are assumed
to have an internal structure, and they are considered as multi-sets or sequences. An interpretation
of a rule is how to transform the premises in the conclusion. Thus the rule 
\[
  \alpha,\beta,\alpha\vdash \gamma
\] 
can be taken to mean that we need two instances of $\alpha$ with an instance of
$\beta$ in between to produce an instance of $\gamma$.
Derivation rules, on the other hand, tell us how to combine rules, to obtain new rules. For example,
the derivation rule \[
\frac{\Gamma\vdash\alpha\qquad \Theta\vdash\alpha\rightarrow\beta}
     {\Gamma,\Theta\vdash\alpha,\alpha\rightarrow\beta,\beta}
\] establishes whether we have a derivation of $\alpha$ from $\Gamma$ and a derivation of
$\alpha\rightarrow\beta$ from $\Theta$, then we can combine the $\Gamma$ and $\Theta$, to obtain a
derivation, where we have $\alpha$ followed by $\alpha\rightarrow\beta$ and then $\beta$. If the
formulas denote activities (or tasks) and resources, then the consequent is a sequence of tasks
describing the activities to be done (and the order in which they have to be executed) to produce an
outcome (and also, what resources are needed). Thus, we can use the rules to model transformation in
a business processes, and derivations as the traces of the process (or the ways in which the process
can be executed or the runs of system).

A formalism that properly models processes should feature some key
characteristics, and one of the most important ones is to identify which
resources are \emph{consumed} after a task has finished its execution.
Consider the notorious vending machine scenario by Girard, where the dollar resource is
spent to \emph{produce} a can of cola. once we get the cola, the
dollar resource is no longer spendable. However, the specifications of a process may include
thousands of rules to represent at their best all the various situations that
may occur during the execution of the process itself: situations where the
information at hand may be incomplete and, sometimes, even contradictory, and
rules encoding possible exceptions. This means that we have to adopt a
formalism that is able to represent and reason with exceptions, and partial
information.

Defeasible Logic (DL) \cite{DBLP:conf/inap/Nute01a} is a non-monotonic rule based formalism, that
has been used to model exceptions and processes. The starting point being that,
while rules define a relation between premises and conclusion, DL takes the stance that multiple relations are possible, and it focuses
on the ``strength'' of the relationships. Three relationships are identified:
\emph{strict rules} specifying that every time the antecedent holds the
consequent hold; \emph{defeasible rules}, when the antecedent holds then,
normally, the consequent holds; and \emph{defeaters} when the antecedent
holds the opposite of the consequent might not hold. An example of rules with
a baseline condition and exception is the scenario the outcome of inserting a
dollar coin in a vending machine is that we get a can of cola, unless the
machine is out of order, or the machine is switched off. Thus, we can
represent this scenario with the rules\footnote{$r_i$ is the name of rule $i$, symbol $\To$ denotes rules meant to derive \emph{defeasible} conclusions, i.e. conclusions which may be defeated by contrary evidence. As it will be clear in Section~\ref{sec:LDL}, defeasible rules, defeaters, and the superiority relation represent the non-monotonic aspects of our framework.}:
\begin{gather*}
r_1: 1\$ \Rightarrow\mathit{cola}\\
r_2: \mathit{OutOfOrder} \Rightarrow \neg\mathit{cola}\\
r_3: \mathit{Off} \Rightarrow \neg\mathit{cola}.
\end{gather*}

In the rest of the paper we shall provide a (brief) introduction to the formalism of linear defeasible logic, 
and then discuss two specific cases in which that fornalism applies to energy management and business processes. In particular
we dedicate Section \ref{sec:desired_properties} to the discussion of the properties introduced in the system
discussed in \cite{DBLP:conf/prima/OlivieriGCBT18}, further detailed in Section \ref{sec:LDL}. Section \ref{sec:applications}
is indeed devoted to the discussion of the applications of Linear Defeasible Logic. Section \ref{sec:concl} discusses some further works and takes conclusions.

\section{Desired properties} % (fold)
\label{sec:desired_properties}

We dedicate this section to detailing which features are needed for 
linear defeasible logic. This section refers to the identified 
properties in \cite{DBLP:conf/prima/OlivieriGCBT18}. 
The properties are summarised in a short way.

\smallskip\noindent\textbf{Ordered list of antecedents.}
% \label{sec:OrderedAnt} 
Given the rule `$r: a, b \To c$', the order in which we derive $a$ and $b$ is
typically irrelevant for the derivation
of $c$. As such, $r$ may indistinctively assume the form `$b, a \To c$'.
Consider a login procedure requiring a username and password.
Whether we insert one credential before the other does not affect a
successful login.

Nonetheless, sometimes it is meaningful to consider an \emph{ordered}
sequence of atoms in the head of a rule, instead of an \emph{unordered} set
of antecedents. 

Naturally, in the same set/list of antecedents, combinations of sequences and
multi-sets of literals is possible. For
instance, ``$r: a; b; (c, d); e\To f$'' represents a situation where, in order
to obtain $f$, we need to first obtain $a$, then $b$, then either $c$ or $d$
in any order, and lastly, only after both $c$ and $d$ are obtained, we need
to obtain $e$. 

The notation `;' is used as a separator in sequences, while `,'
separates in multi-sets.

\smallskip\noindent\textbf{Multi-occurrence/repetitions of literals.}
From these ideas, it follows that some literals may appear
in multiple instances, and that two rules such as ``$r: a; a; b \To c$'' and
``$s: a; b; a \To c$'' are semantically different. Rule $r$ may
describe a scenario
where the order of a product may require two deposit payments followed by a
full payment prior to delivery.

\smallskip\noindent\textbf{Resources consumption.}
% \label{sec:ResCons}
Assume we have two rules, ``$r: a, b \To d$'' and ``$s: a, c \To e$''. If we
are
able to derive $a$, $b$ and
$c$, then $d$
and $e$ are subsequently obtained. Deducing both $d$ and $e$ is a typical
problem of
\emph{resource consumption}. 

In
this
paper, we assume all
literals
to be consumable, since the treatment/derivation of non-consumable literals is
the
same as in Standard DL (SDL). %, since, at it will be clear in the remainder, 

In our framework, deriving a literal means to produce the
corresponding resource. After it has been produced, it can be `consumed' if
it appears in the antecedent of a rule. Naturally, since we have complete
knowledge of our theory, we know exactly in which rules such a
literal appears as an antecedent. Not consuming a resource when such is
available means that either no rule is applied, or that the rule where the
literal appears in as antecedent was \emph{blocked} by a rule for the opposite.

\smallskip\noindent\textbf{Concurrent production.}
% \label{sec:ConcurrentP} 
Symmetrically, we consider two rules with the same conclusion: ``$r: a \To c$
'' and ``$s: b \To c$''. It now seems reasonable that, if both $a$
and $b$ are derived, then we conclude two instances of $c$ (whereas in
classical logics we only know that $c$ holds true). For example, consider a
family where it is tradition to have pizza on Friday evening. Last Friday,
the parents were unable to communicate with each other during the day, and one
baked the pizza while the other bought take-away on the way home.

However, there exists consistent cases where multiple rules for the same
literal produce only
\emph{one} instance of the literal (even if they all fire). For example, both a
digital or handwritten signature would provide permission to proceed with a
request. The same request does not require permission twice: either it is
permitted, or it is not.

\smallskip\noindent\textbf{Resource consumption: A team defeater perspective.}
% \label{sec:ResConsTEAM}
Sceptical logics provide a means to decide which conclusion to draw in case of
contradicting information. A superiority relation is given among
rules for contrary conclusions: it is possible to derive a conclusion only if
there exists a \emph{single} rule stronger than \emph{all} the
rules for the opposite literal.

DL handles conflicts differently, and the idea here is that of
\emph{team defeater}. We do not look at whether there is a single rule
prevailing over all the other rules, but rather whether there exists a
\emph{team} of rules which can jointly defeat the rules for the contrary
conclusion. That is, suppose rules $r'$, $r''$ and $r'''$ all conclude $P$,
whilst $s'$ and $s''$ are for $\neg P$. If $r'>s'$ and $r''>s''$, then the
team defeater made of $\set{r',r''}$ is sufficient to prove $P$.

The focus remains on resource consumption and production. As such, the
questions we need to answer are, again, which resources are consumed, and
how many instances of the conclusion are derived.

Consider the process of writing a scientific publication. If
the paper is accepted, the \emph{manuscript} resource is consumed: it
cannot be submitted again. On the contrary, if the paper is rejected, the
\emph{manuscript} resource is \emph{not} consumed since it can be submitted
again.

\smallskip\noindent\textbf{Multiple conclusions and resource preservation.}
% \label{sec:Multiple}
Consider internet shopping. As soon as we pay for our online order, the bank
account balance decreases, the seller's account increases. Both the seller
and the web site have the shipping address.

The conclusion of a rule is usually a single literal but the above example
suggests that a single rule may produce more than one conclusion. We thus
allow rules to have multiple conclusions, like ``$r: a, b \To c, d$''.

Similar to our discussion 
%in Section~\ref{sec:OrderedAnt}, 
on the ordering of antecedents, we may have any
combination of ordered/unordered sequences of conclusions. In the previous
example, only after we have provided the credit card credentials, our bank
account decreases, whilst we can provide the shipping address before the
credit card credentials, or the other way around.

The notion of multiple conclusions, along with the discussion 
% introduced in Section~\ref{sec:ResConsTEAM}
on team defeaters, leads to another problem. Consider the two
rules ``$r: a \To b;c;d$'' and ``$s: e \To \neg c$'', where no
superiority is given. Do we conclude that $b$ or $d$? What happens if
now we have ``$r: a \To b,c,d$'', and $s$ is stronger then
$r$? Do we conclude that $b$ and $d$ (i.e., only the derivation of $c$ has
been blocked by $s$), or will the production of $b$ and $d$ be affected also?

\smallskip\noindent\textbf{Loops.}
% \label{sec:loops}
The importance of being able to properly handle loops is evident: loops play a
fundamental role in many real life applications, from business processes to
manufacturing. Back to the login procedure, if one of the credentials is
wrong, the process loops back to a previous state, for instance, by asking
the user to re-enter both credentials.

% section loops (end)
% section desired_properties (end)

%\input{SDL}
\section{Language and logical formalisation of RSDL}\label{sec:LDL}

The logic presented here considers consumable literals only: the
formalisation of non-consumable literals is the same of that
in SDL \cite{tocl}, and thus the process would not add any value to
this contribution.
Also we will have either multi-sets or sequences of literals,
but \emph{not} combinations of those.

In linear defeasible logic there are two types of derivations: \emph{strict} and
\emph{defeasible}. \emph{Strict rules} derive indisputable conclusions, i.e.,
conclusions that
which are always true. Thus, if two strict rules have opposite conclusions,
then the resulting logic is inconsistent. On the contrary, defeasible rules
are to derive pieces of information that can be defeated by contrary
evidence. Finally, defeaters are special type of rules whose
purpose is to block contrary evidence: they cannot be used to directly derive
conclusions, but only to prevent other rules to conclude.

In \cite{DBLP:conf/prima/OlivieriGCBT18} it is introduced the language of Resource-driven
Substructural Defeasible Logic (RSDL). $\PROP$ is the set of propositional
atoms, the set $\LIT= \PROP \cup
\set{\neg p | p \in \PROP}$ is the set of literals. The
\emph{complement} of a literal $p$ is denoted by $\non q$; if $q$
is a positive literal $p$, then $\neg q$ is $\neg p$, and if
$q$ is a negative literal $\neg p$, then $\non q$ is $p$.

\begin{definition}\label{def:Rules}

Let $\LAB$ be a set of arbitrary labels. Rules have form ``$r: A(r)
\hookrightarrow C(r)$'':

	\begin{enumerate}
		\item $r\in \LAB$ is a unique name.
\item\label{Alist} $A(r)$ is the \emph{antecedent}, or \emph{body},
of the rule. $A(r)$ can either have the form $A(r)=a_1,\dots,a_n$ to denote a
multi-set, or $A(r)=a_1;\dots;a_n$ to denote a sequence.

\item $\hookrightarrow\in \set{\to, \To, \leadsto}$ denotes a
strict rule, a defeasible rule, and a defeater, respectively.

\item $C(r)$ is the \emph{consequent}, or \emph{head}, of the rule. For the
head, we consider three options: (a) The head is a single literal `$p$', (b)
The
head is a multi-set `$p_1,\dots,p_m$', and (c) The head is a sequence
`$p_1;\dots;p_m$'.
	\end{enumerate}
\end{definition}
Given a set of rules $R$ and a rule $r: A(r) \hookrightarrow C(r)$,
we use
the following abbreviations: (i) $R_s$ is the
subset of strict rules, (ii) $R_{sd}$ is the set of strict and defeasible
rules, (iii) $R[p;i]$ is the set of rules where $p$ appears at
index $i$ in the consequent where the consequent is a sequence, (iv)
$R[p,i]$ when the consequent is a multi-set containing $p$.

\begin{definition}
	A \emph{resource-driven substructural defeasible theory} is %still 
	a tuple $(F, R, \succ)$ where:
(i) $F\subseteq \LIT$ are pieces of information denoting the
resources available at the beginning of the computation. This differs
strikingly from SDL, where facts denote \emph{always-true} statements; (ii)
$R$
is the set of rules; (iii) $\succ$, the superiority relation, is a binary
relation over $R$.
\end{definition}
A theory is \emph{finite} if the sets of facts and rules are finite.
In SDL, a \emph{proof} $P$ of length $n$ is a finite sequence
$P(1),\dots,P(n)$ of \emph{tagged literals} of the type $\pm\Delta
p$, $\pm\partial p$. The idea is that, at every step of the
derivation, a literal is either proven or disproven. 

In Linear Defeasible Logic, we must be
able to derive multiple conclusions in a single
derivation step, and hence we require a mechanism to determine when
premises have been used to derive a conclusion. Accordingly, we modify
the definition of proof to be a matrix.

\begin{definition}\label{def:Proof}
	A proof $P$ in RSDL is a finite matrix
$P(1,1),\dots, P(l,c)$ of \emph{tagged literals} of the type $\pm\Delta p$,
$\pm\partial p$,
$+\sigma p$, $+\Delta p^\checkmark$, and $+\partial p^\checkmark$.

\end{definition}
We assume that facts are
simultaneously true at the beginning of the computation. Notation $+\#
p^\checkmark$, $\#\in\set{\Delta,\partial}$ denotes that $p$ has been consumed.
The distinctive notation for when a
literal is proven and when it is consumed will play a key role to determine
which rules are applicable. The tagged literal $\pm\Delta p$ means that $ p$ is
\emph{strictly
proved/refuted} in $D$, and, symmetrically, $\pm\partial p$ means that
$ p$ is \emph{defeasibly proven/refuted}, $+\sigma p$ indicates that
there are applicable rules for $p$, but the resources for such rules have
already been used.

The set of positive and negative
conclusions is named \emph{extension}. In Standard Defeasible Logic, given a set of facts, a set of
rules, and a superiority
relation, the extension is unique. This is not the case when
resource consumption and ordered sequences are to be taken into account:
depending on the order in which the rules are applied, (rather) different
extensions
can be obtained. In RSDL every distinct derivation
corresponds to an extension.

In SDL, derivations are based on the notions of a rule
being \emph{applicable} or \emph{discarded}. Intuitively, a rule is applicable
when every literal in the antecedent has been proven at a previous step. We
report hereafter the defeasible proof tag in SDL to give the reader a
better understanding of how defeasible conclusions can be drawn.
\begin{tabbing}
If \=$P(n+1) = +\partial p$ then\+\\
  (1) $\exists r\in R_{sd}[p]$: $r$ is applicable and\\
(2) \=$\forall s\in R[\non p]$ either (2.1) $s$ is discarded or (2.2)$\exists
t\in R[p]$: $t$ is applicable and $t\succ s$.
\end{tabbing}
A literal is defeasibly proven when there exists an applicable rule for such
a conclusion and all the rules of the opposite are either discarded, or
defeated by stronger rules. (Strict derivations only differ in that, when a
rule is applicable, we do not care about contrary evidence, and the rule will
always produce its conclusion nonetheless.)

In \cite{DBLP:conf/prima/OlivieriGCBT18} authors proceed incrementally. 
First, they provide definitions of derivability for multi-sets, and further,
definitions for sequences. In both cases,
rules may have only one \emph{single} literal for conclusion. 
\begin{definition}\label{def:ApplicableMS}
A rule $r$ is $\#$\emph{-applicable}, $\#\in\set{\Delta,\partial,\sigma}$, at
$P(l+1,c+1)$ iff
% \begin{enumerate}
% 	\item 
	$\forall$ $a_i\in A(r)$ then $+\# a_i\in P[(1,1)..(l,c)]$.
% \end{enumerate}
Moreover, we say that $r$ is $\#$\emph{-consumable} iff $r$ is $\#$-applicable
and $\exists$ $l'\leq l$ such that $P(l',c)=+\#  a_i$.
\end{definition}
A rule is consumable if it is applicable and, for every literal in its
antecedent, there is an \emph{unused} occurrence. Discardability is the strong
negation of applicability.

\begin{definition}\label{def:DiscardedMS}
A rule $r$ is $\#$\emph{-discarded}, $\#\in\set{\Delta,\partial}$, at
$P(l+1,c+1)$ iff
% \begin{enumerate}
% 	\item 
	$\exists$ $a_i\in A(r)$ such that $-\# a_i\in P[(1,1)..(l,c)].$
% \end{enumerate} 
Moreover, we say that $r$ is $\#$\emph{-non--consumable} iff either $r$ is
discarded, or $\forall$ $l'\leq l$, $P(l',c)\neq +\partial a_i$.
\end{definition}
Lastly, we define the conditions describing when a literal is consumed.

\begin{definition}\label{def:ConsumedLiteral}
Given rule $r$, a literal $a\in A(r)$ is $\#$-consumed,
$\#\in\set{\Delta,\partial}$, at $P(l+1,c+1)$, iff 1. $\exists$ $l'\leq l$ such
that $P(l',c)=+\# a$, and 2. $P(l',c+1)=+\# a^\checkmark$.
\end{definition}
Naturally, two mutually exclusive extensions are possible, based on whether
$+\partial b$ is used by $r_1$ to derive $+\partial c$, or by $r_2$ to derive
$+\partial d$. 

\subsection*{Proof tags for multi-sets in the antecedent and single conclusion}
% (fold)
\label{sec:proof_tags_for_multi_sets_in_the_antecedent_and_single_conlusion}

% section proof_tags_for_multi_sets_in_the_antecedent_and_single_conlusion (end)

We begin with: (1) the antecedent is a
multi-set, (2) single literal in the conclusion.
\begin{tabbing}
 $+\Delta$: \= If $P(l+1,c+1)=+\Delta p$ then\+\\
 (1) \=$p\in F$, or  (2) $\exists r\in R_s[p]$ s.t. $r$ is
$\Delta$-consumable and $\forall
a_j\in A(r)$, $a_j$ is $\Delta$-consumed.
 %\> (2.1) , and\\
 %\> (2.2) $P(l_i,c_i) = +\Delta a_i^\checkmark$.
% \> (2.3) $n_1 < n_2 < \dots < n_m$.
	\end{tabbing}
Literal $q$ is definitely provable if either is a fact, or there is a strict
consumable rule for $p$. Condition (2) actually consumes the literals by
replacing $+\Delta a_j$ with $+\Delta a_j^\checkmark$.
\begin{tabbing}
 $-\Delta$: \= If $P(l+1,c+1)=-\Delta p$ then\+\\
(1) \=$p\notin \FACTS$ and (2) $\forall r\in R_s[p]$, $r$ is
$\Delta$-non-consumable.

\end{tabbing}
Literal $q$ is definitely refuted if $p$ is not a fact, and every
rule for $p$ is non-consumable.
\begin{tabbing}
 $+\partial$: \= If $P(l+1,c+1)=+\partial  p$, then\+\\
 (1) \=$+\Delta  p\in P(l,c)$ or\\
 (2) \= (1) $-\Delta\non  p \in P(l,c)$ and\+\\
     (2) \= $\exists r\in R_{sd}[p]$ $\partial$-consumable and (3) $\forall s \in R[\non p]$ either\+\\
 	     (1) \= $s$ is $\partial$-discarded, or\\
	     (2) $\exists t\in R[p]$ $\partial$-consumable, $t\succ s$, and\\
(3) if $\exists w\in R[\non p]$ $\partial$-applicable, $t\succ w$, then\+\\
(1) \=$\forall a_j\in A(t)$, $a_j$ is $\partial$-consumed, otherwise  (2) $\forall a_k\in A(r)$, $a_k$ is $\partial$-consumed.
\end{tabbing}
Condition (1) is to inherit a defeasible derivation from a definite
one. Condition (2.1) ensures that the logic is sound. Condition
(2.2) requires that there is a rule $r$ that is triggered by
literals that: (i) have been previously proven, (ii) have not yet been
consumed. Clause (2.3.1) is that, to rebut
an attacking argument, either it is possible to show that some of its premises have
been refuted, or is defeated by stronger, consumable rules. The final part is
to determine which resources
are consumed during the derivation of $p$. This variant assumes that
only the rules in the winning team defeater consume resources, whilst the
`defeated rules' do not.

For $-\partial$ authors of \cite{DBLP:conf/prima/OlivieriGCBT18} use the strategy similar to that used in \cite{ecai2000-5}
to provide proof conditions for the ambiguity propagating variant of SDL, that
is, we make it easier to attack a rule (2.2.2). Also, derivations literals tagged with $-\partial$ do not
consume of resources, in line with what above.
\begin{tabbing}
$-\partial$: \= If $P(l+1,c+1)=-\partial p$, then \+\\
  (1) $-\Delta p\in P[l,c]$ and\\
  (2)\=(1) $+\Delta\non p\in P[l,c]$ or\+\\
     (2) \=$\forall r\in R_{sd}[ p]$ either\+\\
        (1)\= $r$ is $\partial$-discarded or (2) $\exists s\in R[\non p]$ s.t.\+\\
            (1)\= $s$ is $\sigma$-applicable and
            (2) $\forall t\in R[ p]$ either
                 $t$ is $\partial$-discarded or
                 not $t\succ s$.  
\end{tabbing}
The idea behind $+\sigma$ is that there are applicable, non-defeated rules for
the consequent, irrespective whether the premises have been used or not.
\begin{tabbing}
$+\sigma$: \=If $P(l+1,c+1)=+\sigma p$ then\+\\
 (1) \= $\Delta p\in P[l,c]$ or\\ 
 (2) $\exists r\in R_{sd}[p]$ s.t. (1)   $r$ is $\sigma$-applicable and
     (2) $\forall s\in R[\non p]$ either
          $s$ is $\partial$-discarded or $s\not\succ r$.
\end{tabbing}
Assume that, at $P(4,4)$, $r_0$ is taken into
consideration; $r_0$ is consumable, but so is $r_3$, and no
superiority is given between $r_0$ and $r_3$. Actually, $r_2$ is
consumable and stronger than $r_3$. Accordingly, the team
defeater allows us to prove $+\partial d$, and only $b$ is consumed
in this process. Thus $a$ is still available,
and can be used at $P(5,5)$ to get $+\partial e$, via
$r_1$. Note that we do not consume resource $+\Delta c$, since $r_2\succ r_3$.

\subsection*{Proof tags for sequences in the antecedent and single conclusion}
% (fold)
\label{sec:proof_tags_for_sequences_in_the_antecedent_and_single_conlusion}

% section proof_tags_for_sequences_in_the_antecedent_and_single_conlusion (end)

When considering sequences in the antecedent, definitions of applicable,
discarded, and consumable must be revised. A rule is \emph{sequence applicable}
when the derivation order reflects the
order in which the literals appear in the antecedent,

\begin{definition}\label{def:SeqAppl}
A rule $r\in R[p]$ is $\#$-\emph{sequence applicable},
$\#\in\set{\Delta,\partial}$, at $P(l+1,c+1)$ iff for all $a_i\in A(r)$:
	
\begin{enumerate}

\item\label{cond:Appl1} there exists $c_i \leq c$ such that $P(l_i,c_i) = +\#
a_i$, $l_i \leq l$, and
	
	\item for all $a_j\in A(r)$ such that $i < j$, then
	
\item\label{cond:Appl3} for all $c_j \leq c$ such that $P(l_j,c_j) = +\# a_j$,
$l_j \leq l$ then $l_i < l_j$ and $c_i < c_j$.

\end{enumerate}

We say that $r$ is $\#$-\emph{sequence consumable} iff is
$\#$-\emph{applicable} and 4. $P(l_i,c)=+\# a_i$.
\end{definition}
A rule is sequence discarded when there exists a literal in the antecedent,
 which has been previously disproven, or there are two proven literals in
the antecedent, say $a$ and $b$, such that $a$ appears before
$b$, and one proof for $b$ is before every proof for $a$.

\begin{definition}\label{def:SeqDisc}
A rule $r\in R[p]$ is $\#$-\emph{sequence discarded},
$\#\in\set{\Delta,\partial}$, at $P(l+1,c+1)$ iff for there exists $a_i\in
A(r)$ such that either
	
\begin{enumerate}

	\item $-\#a_i \in P(l-1,c-1)$, or 
	
	\item for all $c_i \leq c$ such that $P(l_j,c_i)=+\# a_i$, $l_i \leq l$, then
	
	\begin{itemize}
		\item there exists $a_j\in A(r)$, with $i < j$, such that
		
\item there exists $c_j \leq c$ such that $P(l_j, c_j)=+\# a_j$, $l_j \leq l$,
and $c_i > c_j$, $l_i > l_j$.
	\end{itemize}

\end{enumerate}
\end{definition}
The definition of being $\#$-consumed remains the same as before. The
proof tags for strict and defeasible conclusions with (i)
sequences in the antecedent, and (ii) a single conclusion can be obtained by
simply replacing: (a) $\#$-applicable with $\#$-sequence applicable, (b)
$\#$-consumable with $\#$-sequence consumable, and (c) $\#$-discarded with
$\#$-sequence discarded.	
Assume the rules are activated in this order: first $r_1$, then $r_2$, last
$r_3$. Thus, $P(4,4)=+\partial a$, $P(5,5)=+\partial a$, and
$P(6,6)=+\partial b$. The derivation order between $b$ and the second
occurrence of $a$ has not been complied with, and $r_0$ is sequence
discarded. Same if the order is `$r_3$, $r_1$, $r_2$', whilst `$r_2$, $r_3$,
$r_1$' is a legit order to let $r_0$ be sequence applicable.

\subsection*{Proof tags for sequences in both the antecedent and conclusion}
\label{sec:proof_tags_for_sequences_in_both_the_antecedent_and_conclusion}

Even when we consider sequences in the consequent, a literal's strict
provability or refutability depends only upon whether the strict rule (where the
literal occurs) is sequence consumable or not. As such, given a strict rule
$r\in R_s[p;j]$, still $p$'s strict
provability/refutability depends only upon whether $r$ is strictly sequence
consumable or not. However, now we also have to verify that, if $r\in
R_s[q;j-1]$, we prove $p$ immediately after $q$.
The resulting new formalisations of $+\Delta$ and $+\partial$ are as follows
(negative proof tags are trivial and thus omitted):

\begin{tabbing}
 $+\Delta$: \= If $P(l+1,c+1)=+\Delta p$ then\+\\
 (1) \=$p\in F$, or (2) (1) $\exists r\in R_s[p;j]$ s.t. (2) $r$ is
$\Delta$-sequence-consumable,\\
 \> (3) $r\in R_s[q;j-1]$ and $P(l+i-1,c+1)=+\Delta q$, (4) $\forall a_j\in
A(r)$, $a_j$ is $\Delta$-consumed.
 %\> (2.1) , and\\
 %\> (2.2) $P(l_i,c_i) = +\Delta a_i^\checkmark$.
% \> (2.3) $n_1 < n_2 < \dots < n_m$.
\end{tabbing}

\begin{tabbing}
 $+\partial$: \= If $P(l+i,c+1)=+\partial p$, then\+\\
 (1) \=$+\Delta p\in P(l,c)$ or (2) (1) $-\Delta\non p \in P(l,c)$ and\+\\
     (2) \= (1) $\exists r\in R_{sd}[p;j]$ $\partial$-sequence-consumable and\\
	 	\> (2) $\exists r\in R[q;j-1]$ and (3) $P(l+i-1,c+1)=+\partial q$, and\\
     (3) \= $\forall s \in R[\non p]$ either\+\\
 	     (1) \= $s$ is $\partial$-sequence-discarded, or\\
	     (2) $\exists t\in R[p]$ $\partial$-sequence-consumable, $t\succ s$, and\\
(3) if $\exists w\in R[\non p]$ $\partial$-sequence-applicable, $t\succ w$,
then\+\\
(1) \=$\forall a_j\in A(t)$, $a_j$ is $\partial$-consumed, otherwise (2)
$\forall a_k\in A(r)$, $a_k$ is $\partial$-consumed.
\end{tabbing} 

% section proof_tags_for_sequences_in_both_the_antecedent_and_conclusion (end)

\subsection*{Proof tags for sequences in the antecedent and multi-sets in the
conclusion} % (fold)
\label{sec:proof_tags_for_sequences_in_the_antecedent_and_multi_sets_in_the
_conclusion}
 
We now consider multi-sets in the conclusion. Strict provability does not
change with respect to the one described in the
previous section, and is therefore omitted. When considering a `team defeater
fight', two variants are possible. In this first variant, we draw a
conclusion only if there is a winning team defeater for each literal in the
conclusion.
\begin{tabbing}
 $+\partial$: \= If $P(l+i,c+1)=+\partial p$, then\+\\
 (1) \=$+\Delta p\in P(l,c)$ or (2) (1) $-\Delta\non p \in P(l,c)$ and\+\\
     (2) $\exists r\in R_{sd}[p,j]$ $\partial$-sequence-consumable and\\
     (3) \= $\forall s \in R[\non  q]$ such that $ q\in C(r)$ either (1) $s$ is
$\partial$-sequence-discarded, or\+\\
(2) \= $\exists t\in R[ q]$ $\partial$-sequence-consumable, $t\succ s$, and\\
(3) if $\exists w\in R[\non p]$ $\partial$-sequence-applicable, $t\succ w$,
then\+\\
(1) \=$\forall a_j\in A(t)$, $a_j$ is $\partial$-consumed, otherwise (2)
$\forall a_k\in A(r)$, $a_k$ is $\partial$-consumed.
\end{tabbing}

In this latter variant, we limit the comparison on
the individual literal.
\begin{tabbing}
 $+\partial$: \= If $P(l+i,c+1)=+\partial p$, then\+\\
 (1) \=$+\Delta p\in P(l,c)$ or\\
 (2) \= (1) $-\Delta\non p \in P(l,c)$ and\+\\
     (2) $\exists r\in R_{sd}[p,j]$ $\partial$-sequence-consumable and\\ \=(3)
$\forall s \in R[\non p]$ either\+\\
 	     (1) \= $s$ is $\partial$-sequence-discarded, or\\
	     (2) $\exists t\in R[p]$ $\partial$-sequence-consumable, $t\succ s$, and\\
(3) if $\exists w\in R[\non p]$ $\partial$-sequence-applicable, $t\succ w$,
then\+\\
(1) \=$\forall a_j\in A(t)$, $a_j$ is $\partial$-consumed, otherwise \\(2)
$\forall a_k\in A(r)$, $a_k$ is $\partial$-consumed.
\end{tabbing} 
%

% section

%\input{Considerations}
\section{Applications of linear defeasible logic} % (fold)
\label{sec:applications}
This section is devoted to the description of two application scenarios for the combination of resource consumption and 
exceptions into a logical framework. In both scenarios, the basic idea of the application is to provide an abstract 
formalisation of a context. In the first case scenario, we shall discuss how a reasoning process can provide decision onto
a set of rules that connect energy resources in a defeasible way. In the second case scenario we discuss two aspects of
business process management that deserve linear defeasible logic treatment.

In both cases, the discriminant for choosing a logical framework sites on the reasoning capability of
such frameworks. Ideally, we can specify the rules that govern a context and compute a solution to these
rules, that consist, in one case, of a plan of energy resource production and consumption, and
in the other case, of a process description.

Description of the application scenarios is made entirely in an informal way, for the sake of
conciseness. Formalization is left to further work.

\subsection{Energy management}
A top-level description of an energy production and consumption context is defined by a set of rules and facts, where 
rules are employed to describe defeasible transformations of energy, and facts represent energy resources.
We provide here three basic cases of application

\begin{scenario}
	\textbf{Energy saving}\\
	One of the most difficult concepts to deal with in the definition of methods for energy saving is the definition of the comparison of processes (the so called
	\emph{consumption profiles}, in order to detect the best solution. Consider a classical home configuration, as represented in Figure \ref{fig:elettrascenario}.
	
	\begin{figure}[h!]
		\centering
			\includegraphics[scale=0.25]{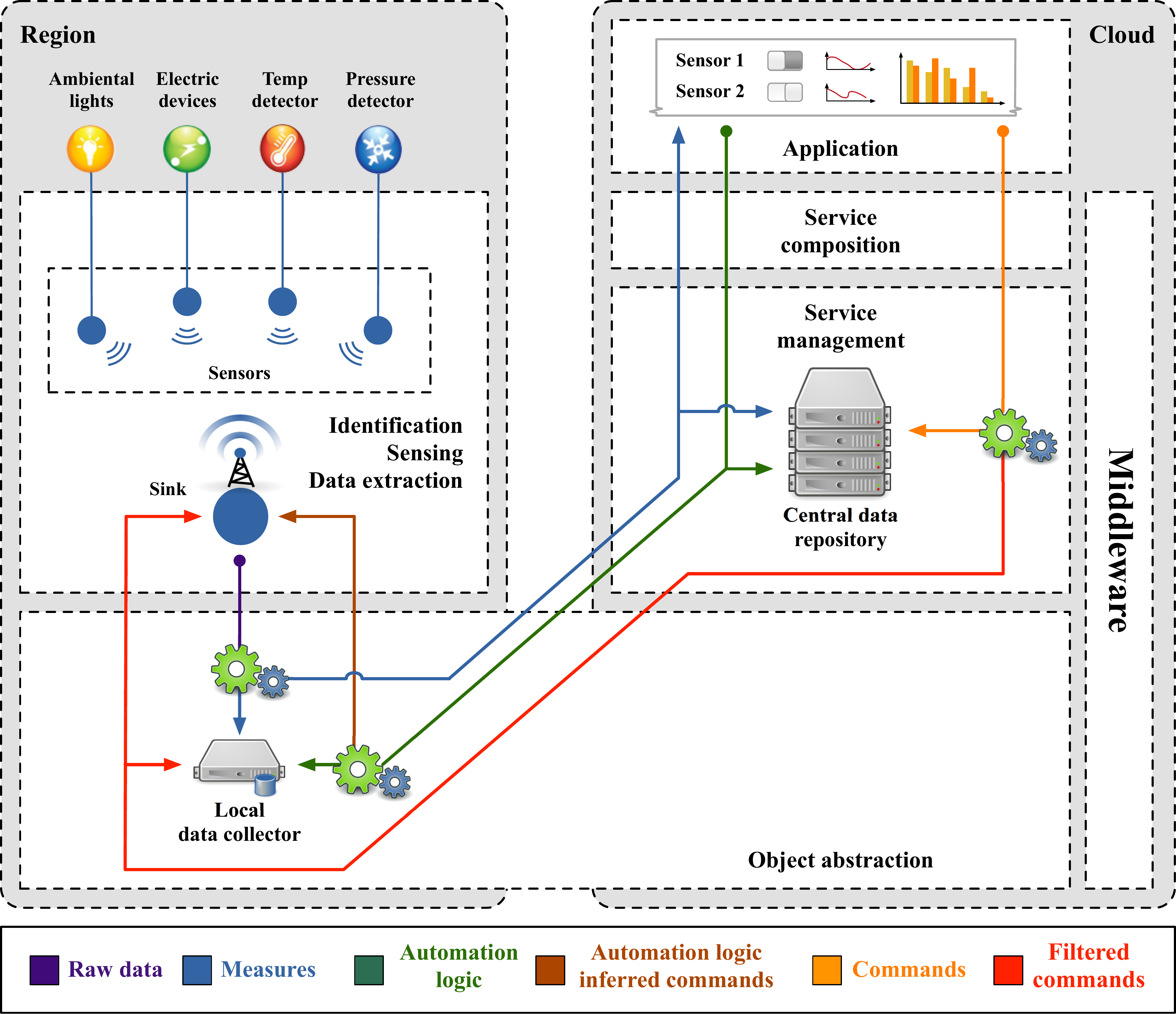}
		\caption{The architecture of \emph{MyElettra} an ambient intelligence technology for energy efficiency.}
		\label{fig:elettrascenario}
	\end{figure}
	
	The most common approach is \emph{ranking}, that is based upon the usage of multiplicative or additive measures attached to the devices, able to represent the \emph{typical}
	concumption, usually, the \emph{average} estimated consumption. This approaches have shown to be very inaccurate in forecasting the behaviour, as numerous \emph{exceptions} can
	significantly change the resulting value of energy, for the local measure used typically is \emph{absorbed power}, and energy is instead the integral of power. If the sampling
	processes that has generated the above defined scenario, has valued the absorbed power in instants of time where the measure has lowered suddenly, then the
	result would be strongly inaccurate.
	
	A more sophisticated approach consists in using \emph{ambient intelligence} data. Again this configuration suffer from two drawbacks: the limited amount of data that
	can be collected during the decision process can be strongly influential on the computer solution, and reactivity to the environment behaviours can be quite useless
	in energy saving for a number of devices whose consumption curve is very inefficient if compressed in short intervals (for instance, electric heating is very inefficient
	if it is turned off and on frequently, therefore is a very bad candidate to the usage of methods based on presence sensors).
	
	Towards a more logical viewpoint support to the decision can be obtained by using defeasible logic, as shown in particular in \cite{Cristani2016885,Tomazzoli2017345}. 
	Starting from that concept, consisting in the usage of defeasible methods to represent different, possibly conflicting rules to save energy in home, home-office, small
	office contexts, we can apply the concept of linear defeasible logic to analogous contexts, enhancing the representation with a novel correspondence.\\ \ \\
	
	\noindent \textbf{Using LDL} We map into rules the behaviour that we desire for a compenent of an energy-absorbing system, that start how those confiurations
	produce the effects that we desire for them, and map into exceptions the \emph{energy wastes}. In other terms, we assume that a rule \emph{saves energy} in the sense that
	it achieves a specific goal of functionality of the system used to control, by using a certain amount of energy. An exception, while possibly stopping achievement of the goal
	(as in any common planning scenario with exceptions, as represented, for instance, in \cite{Governatori2011282}) can be specifically use to represent
	the occasional possibility that the \emph{energy consumption} needed for that particular functionality is higher than usual. That method will be able to prevent
	both drawbacks of the ambient intelligence and rule-based opposite approaches. In the first case, in fact, we have seen employed mainly statistical methods, whilst in the 
	second approach the specification of native rules can provide a good accommodation for common sense rationalizations but no optimization issues can be applied.
	
	If we compute the \emph{least consuming scenario} but goal compliant, we obtain a system that reduces wastes, as modeled by the designer of the ambient intelligent system,
	a neat novelty in the context of energy saving.
 \end{scenario}

\begin{scenario}
	\textbf{Energy production waste control}\\
	Analogously to the scenario of energy saving, we can represent the production of energy with linear defeasible rules, where the goal of a rule is twofold:
	
	\begin{itemize}
		\item Representing the desired production;
		\item Providing room for the \emph{disperded} energy, in the production process.
	\end{itemize}
	
	Once we have specificed correctly the rules of such a production system, the optimization techniques that classically apply to plans (see, for instance, \cite{Tambe201517,Burggraeve2017134} for recent advancements) can generate perfect plans to execute. However, in this configurations, the practical cases
	should identify exceptional configurations in which it is not possible to forecast in a correct way the energy absorbed (or other numerical variables associated to the 
	plan, such as delays of time, or money waste). The representation of this exception in an explicit way can be very useful to generate correct plans.
\end{scenario}

\subsection{Modelling business processess}

\begin{scenario}
	\textbf{Business process compliance by design or revision}\\
	 Long line of application of defeasible logic to business process compliance can be found in literature, mostly cited in Section \ref{section:intro}. In particular we 
	  refer here to \cite{DBLP:conf/prima/OlivieriCG15} where compliance is used in a planning context. What is precisely the business process compliance problem, and why should we consider
	  applying Linear Defeasibly Logic to it?
	  
	  A business process is said to be compliant when it achieves the goals it has been conceived for in a way that does not violate the norms in the normative
	  background the business process has to be executed. For instance, a procedure to build a safety device in a car, say the safety belts, can be respectful of the
	  norms of the country it is produced in while workers wear adequate shoes, the environment of the production is clean enough and so on. One of the most common 
	  problems with this issue is to provide algorithms that allow to build compliant business process \emph{while designing the process} and not \emph{a posteriori}. In 
	  the second undesired scenario, in fact, there will be a risk of generating an uncompliant process, and therefore making it fail. Instead it will definitely make sense
	  to provide \emph{only} compliant processes. Methods used so far produce processes with the desired properties but do not take into consideration resource consumption.
	  
	  After having provided a configration issue as stated above, and in line to the discussion of Section \ref{section:intro}, we may notice that the major problem in these
	  configuration sites on the \emph{exception types} that can be introduced. In a pleminary study, related to MES applications (where the cost accounting is relevant) 
	  we have found that numerous exceptions have not been considered in the generation of business processes, and the combination of these have definite effects on the 
	  compliance itself. A partial yet incomplete list is below:
	  
	  \begin{itemize}
	  	\item \textbf{Safety exceptions} occur when a process need to be interrupted for an accident occured.
		\item \textbf{Stock break exceptions} occur when the feedstock of a machinery ends out.
		\item \textbf{Lack of energy} produces stops in the production process.
		\item \textbf{Strikes, protests, or other interruptions} produce stop in the process.
	  \end{itemize}

	  Methods to reason about these exceptions cannot be safely based upon the usage of pure defeasible exceptions. It is rather obvious that all the above mentioned
	  cases involve \emph{resource consumptions} and should therefore treated in the way presented in \cite{DBLP:conf/prima/OlivieriGCBT18} and discussed here as an application
	  issue.
	
	Business process revisions have been studied widely. A very specific study we refer to here is \cite{Scannapieco2013324} where revision is studied in terms of 
	preserving compliance when changing goals or normative background. Analogous to what has been provided above can occur when one configuration has been chosen and
	we aim at revising the choice for improvements.
	
\end{scenario}

\section{Conclusions and related work}\label{sec:concl}

Some of the authors have already carried out an effort in the direction of applying defeasible logic to energy management, in \cite{Cristani2016885,Tomazzoli2017345}. 
This complemented the investigations on applying logic and machine learning to energy management \cite{Cristani2015324,Cristani2014157,Tomazzoli2018164}. Analogously
some authors applied defeasible logic to business processes \cite{Governatori201399,Governatori2011282} following the long line of investigations cited in the introduction
\cite{Kanovich199748,Tanabe1997156,DBLP:journals/apal/EngbergW97,rao2006composition,DBLP:conf/prima/OlivieriGSC13,DBLP:conf/prima/OlivieriCG15,DBLP:conf/edoc/GhooshchiBGOS17}.

The novelty introduced in \cite{DBLP:conf/prima/OlivieriGCBT18} and discussed in this paper has no precedent for business processes for the combination itself being
new as a whole. Conversely, usage of linear logic to model business processes is new for itself. There have been some investigations trying to determine applicability
of linear logic to planning and to the modelling of complex processes, for Petri nets (see \cite{Pradin-Chézalviel2010481} for a relatively recent review, and 
\cite{Demeterová201691} for more recent advacements).

We discussed the applications of a recently introduced logical apparatus that deals with the problem of
manipulating resource consumption in non-monotonic reasoning. We illustrated in particular
applications to energy management and business processes.

Applications of linear logic to problems indirectly related to business processes such as Petri Nets can be found in 
\cite{Kanovich199748,Tanabe1997156,DBLP:journals/apal/EngbergW97}. However,
such approaches are not able to handle in a natural fashion the aspect of
exceptions. In \cite{linear-agent,linear-agent-Mel1,linear-agent-Mel2}, the
authors propose the use of Linear Logic to generate which plans the agent
adopts to achieve its goals. In the same spirit, \cite{tplp:goal}
address the problem of agents being able to take decisions from partial,
incomplete, and possibly inconsistent knowledge bases.
\bibliographystyle{eptcs}
\bibliography{thisbiblio}

%\appendix
%\section{Formal proofs}
%\input{FormalProofs}

\end{document}